\documentclass{article}

\usepackage{natbib}
\usepackage{authblk}
\usepackage{fullpage}

\usepackage[utf8]{inputenc} %
\usepackage[T1]{fontenc}    %
\usepackage{hyperref}       %
\usepackage{url}            %
\usepackage{booktabs}       %
\usepackage{amsfonts}       %
\usepackage{nicefrac}       %
\usepackage{microtype}      %
\usepackage{xcolor}         %

\title{Learning Barrier Certificates: Towards Safe Reinforcement Learning with Zero Training-time Violations}
\usepackage{graphicx}
\usepackage{amsmath}
\usepackage{amsthm}
\usepackage{dsfont}
\usepackage{cleveref}
\usepackage{enumitem}
\usepackage{algorithm}
\usepackage[noend]{algorithmic}
\usepackage{caption}
\usepackage{subcaption}
\usepackage{wrapfig}
\newcommand{\algo}{CRABS}

\newcommand{\RR}{\mathds{R}}

\newcommand{\policy}{\pi}
\newcommand{\piexpl}{\policy^\mathrm{expl}}
\newcommand{\piinit}{\policy_\mathrm{init}}

\newcommand{\buffer}{\widehat{D}}
\newcommand{\model}{\widehat{\mathcal{T}}}

\newcommand{\return}{J}
\newcommand{\States}{\mathcal{S}}
\newcommand{\Actions}{\mathcal{A}}

\newcommand{\Ssafe}{\mathcal{S}_\text{safe}}
\newcommand{\Sunsafe}{\mathcal{S}_\text{unsafe}}

\newcommand{\reward}{r}
\newcommand{\Lya}{h}

\newcommand{\truemodel}{T}

\newcommand{\Lparam}{\phi}
\newcommand{\Piparam}{\theta}
\newcommand{\Mparam}{\omega}
\newcommand{\Qparam}{\psi}

\newcommand{\level}{\mathcal{C}}
\newcommand{\levelset}{\mathcal{C}_{h_\Lparam}}
\newcommand{\barrier}{\mathcal{B}_{\textup{unsafe}}}
\newcommand{\Binit}{\mathcal{B}_{\textup{init}}}
\newcommand{\grad}{\nabla}

\newcommand{\batch}{\mathcal{B}}
\newcommand{\gLc}{barrier certificate }

\newcommand{\U}{U}
\newcommand{\diag}{\text{diag}}
\newcommand{\env}[1]{\textbf{\emph{#1}}}
\newcommand{\Ch}{\mathcal{C}_h}
\newcommand{\backuppolicy}{\pi^\text{safeguard}}

\def\shownotes{1}  %
\ifnum\shownotes=1
\newcommand{\authnote}[2]{[#1: #2]}
\else
\newcommand{\authnote}[2]{}
\fi

\newtheorem{definition}{Definition}

\author[1]{Yuping Luo\thanks{yupingl@cs.princeton.edu}}
\author[2]{Tengyu Ma\thanks{tengyuma@stanford.edu}}
\affil[1]{\normalsize Computer Science Department, Princeton University}
\affil[2]{\normalsize Department of Computer Science and Statistics, Stanford University}

\begin{document}

\maketitle

\begin{abstract}

Training-time safety violations have been a major concern when we deploy reinforcement learning algorithms in the real world.
This paper explores the possibility of safe RL algorithms with \textit{zero} training-time safety violations in the challenging setting where we are only given a safe but trivial-reward initial policy without any prior knowledge of the dynamics model and additional offline data.
We propose an algorithm, \textbf{C}o-t\textbf{ra}ined \textbf{B}arrier Certificate for \textbf{S}afe RL (CRABS),which iteratively \textit{learns} barrier certificates, dynamics models, and policies. The barrier certificates, learned via adversarial training, ensure the policy's safety assuming calibrated learned dynamics model. We also add a regularization term to encourage larger certified regions to enable better exploration.
Empirical simulations show that zero safety violations are already challenging for a suite of simple environments with only 2-4 dimensional state space, especially if high-reward policies have to visit regions near the safety boundary.  Prior methods require hundreds of violations to achieve decent rewards on these tasks, whereas our proposed algorithms incur zero violations. \footnote{The source code of this work is available at \url{https://github.com/roosephu/crabs}. }%
\end{abstract}

\section{Introduction}
\renewcommand{\cite}{\citep}
\label{sec:intro}
Researchers have demonstrated that reinforcement learning (RL) can solve complex tasks such as Atari games \cite{mnih2015human}, Go \cite{silver2017mastering}, dexterous manipulation tasks~\cite{akkaya2019solving}, and many more robotics tasks in simulated environments~\cite{haarnoja2018soft}.
However, deploying RL algorithms to real-world problems still faces the hurdle that they require many unsafe environment interactions.
For example, a robot's unsafe environment interactions include falling and hitting other objects, which incur physical damage costly to repair.
Many recent deep RL works reduce the number of environment interactions significantly (e.g., see~\citet{haarnoja2018soft,fujimoto2018addressing,janner2019trust,dong2020expressivity,luo2019algorithmic,chua2018deep} and reference therein), but the number of unsafe interactions is still prohibitive for safety-critical applications such as robotics, medicine, or autonomous vehicles~\citep{berkenkamp2017safe}.

Reducing the number of safety violations may not be sufficient for these safety-critical applications---we may have to eliminate them.
This paper explores the possibility of safe RL algorithms with \textit{zero safety violations} in both training time and test time. We also consider the challenging setting where we are only given a safe but trivial-reward initial policy.

A recent line of works on safe RL design novel actor-critic based algorithms under the constrained policy optimization formulation~\citep{thananjeyan2021recovery,srinivasan2020learning,bharadhwaj2020conservative,yang2020accelerating,stooke2020responsive}.  They significantly reduce the number of training-time safety violations.
However, these algorithms fundamentally learn the safety constraints by contrasting the safe and unsafe trajectories.
In other words, because the safety set is only specified through the safety costs that are observed \textit{postmortem}, the algorithms only learn the concept of safety through seeing unsafe trajectories.
Therefore, these algorithms cannot achieve zero training-time violations. For example,  even for the simple 2D inverted pendulum environment, these methods still require at least 80 unsafe trajectories (see \Cref{fig:main-results} in \Cref{sec:experiment}).

Another line of work utilizes ideas from control theory and model-based approach~\citep{cheng2019end,berkenkamp2017safe,taylor2019episodic,zeng2020safety}. These works propose sufficient conditions involving certain Lyapunov functions or control barrier functions that can certify the safety of a subset of states or policies~\citep{cheng2019end}.
These conditions assume access to calibrated dynamical models. They can, in principle, permit safety guarantees without visiting any unsafe states because, with the calibrated dynamics model, we can foresee future danger. However, control barrier functions are often non-trivially \textit{handcrafted} with prior knowledge of the environments~\citep{ames2019control,nguyen2016exponential}.

This work aims to design model-based safe RL algorithms that empirically achieve zero training-time safety violations by \textit{learning} the barrier certificates \textit{iteratively}.
We present the algorithm \textbf{C}o-t\textbf{ra}ined \textbf{B}arrier Certificate for \textbf{S}afe RL (CRABS),  which alternates between \textit{learning} barrier certificates that certify the safety of \textit{larger} regions of states, optimizing the policy, collecting more data within the certified states, and refining the learned dynamics model with data.\footnote{We note that our goal is not to provide end-to-end formal guarantees of safety, which might be extremely challenging---nonconvex minimax optimizations and uncertainty quantification for neural networks are used as sub-procedures, and it's challenging to have worst-case guarantees for them.}

The work of~\citet{richards2018lyapunov} is a closely related prior result, which learns a Lyapunov function given a fixed dynamics model via discretization of the state space.
Our work significantly extends it with three algorithmic innovations. First, we use adversarial training to learn the certificates, which avoids discretizing state space and can potentially work with higher dimensional state space than the two-dimensional problems in~\citet{richards2018lyapunov}.
Second, we do not assume a given, globally accurate dynamics model; instead, we learn the dynamics model from safe explorations. We achieve this by co-learning the certificates, dynamics model, and policy to iteratively grow the certified region and improve the dynamics model and still maintain zero violations. Thirdly, the work~\citet{richards2018lyapunov} only certifies the safety of some states and does not involve learning a policy. In contrast, our work learns a policy and tailors the certificates to the learned policies.
In particular, our certificates aim to certify only states near the trajectories of the current and past policies---this allows us to not waste the expressive power of the certificate parameterization on irrelevant low-reward states.

We evaluate our algorithms on a suite of tasks, including a few where achieving high rewards requires careful exploration near the safety boundary. For example, in the \env{Swing} environment, the goal is to swing a rod with the largest possible angle under the safety constraints that the angle is less than 90$^\circ$. We show that our method reduces the number of safety violations from several hundred to zero on these tasks.

\section{Setup and Preliminaries}
\subsection{Problem Setup}

We consider the standard RL setup with an infinite-horizon \textit{deterministic} Markov decision process (MDP).
An MDP is specified by a tuple $(\States, \Actions, \gamma, \reward, \mu, \truemodel)$, where $\States$ is the state space, $\Actions$ is the action space, $\reward: \States \times \Actions \to \RR$ is the reward function, $0 \leq \gamma < 1$ is the discount factor, $\mu$ is the distribution of the initial state,
and $\truemodel: \States \times \Actions \to \States$ is the deterministic dynamics model.
Let $\Delta(\mathcal{X})$ denote the family of distributions over a set $\mathcal{X}$.
The expected discounted total reward of a policy $\pi: \States \to \Delta(\Actions)$ is defined as
$$
\return(\pi) = \mathbb{E} \left[ \sum_{i=0}^\infty \gamma^i r(s_i, a_i) \right],
$$
where $s_0 \sim \mu, a_i \sim \pi(s_i), s_{i+1}=\truemodel(s_i, a_i)$ for $i \geq 0$. The goal is to find a policy $\policy$ which maximizes $\return(\policy)$.

Let $\Sunsafe\subset \States$ be the set of unsafe states specified by the user.
The user-specified safe set $\Ssafe$ is defined as $\States \backslash \Sunsafe$.  %
A state $s$ is (user-specified) safe if $s \in \Ssafe$. A trajectory is safe if and only if all the states in the trajectory are safe.
An initial state drawn from $\mu$ is assumed to safe with probability 1.
We say a deterministic policy $\policy$ is safe starting from state $s$, if the infinite-horizon trajectory obtained by executing $\policy$ starting from $s$ is safe.
We also say a policy $\policy$ is safe if it is safe starting from an initial state drawn from $\mu$ with probability 1.
A major challenge toward safe RL is the existence of irrecoverable states which are currently safe but will eventually lead to unsafe states regardless of future actions. We define the notion formally as follows.
\begin{definition}
    A state $s$ is \emph{viable} iff there exists a policy $\policy$ such that $\policy$ is safe starting from $s$, that is, executing $\pi$ starting from $s$ for infinite steps never leads to an unsafe state.  A user-specified safe state that is not viable is called an \emph{irrecoverable} state.
\end{definition}
We remark that unlike \citet{srinivasan2020learning,roderick2020provably}, we do not assume all safe states are viable. We rely on the extrapolation and calibration of the dynamics model to foresee risks. A calibrated dynamics model $\model$ predicts a confidence region of states $\model(s, a) \subseteq \States$, such that for any state $s$ and action $a$, we have $\truemodel(s, a) \in \model(s, a)$.

\subsection{Preliminaries on Barrier Certificate}
\label{sec:prelim-barrier}

Barrier certificates are powerful tools to certify the stability of a dynamical system. Barrier certificates are often applied to a continuous-time dynamical system, but here we describe its discrete-time version where our work is based upon. We refer the readers to \citet{prajna2004safety,prajna2005necessity} for more information about continuous-time barrier certificates.

Given a discrete-time dynamical system $s_{t+1} = f(s_t)$ \textit{without control} starting from $s_0$, a function $h: \States \to \RR$ is a barrier certifcate if for any $s \in \States$ such that $h(s) \geq 0$, $h(f(s)) \geq 0$. \citet{zeng2020safety} considers a more restrictive requirement: For any state $s \in \States$, $h(f(s)) \geq \alpha h(s)$ for a constant $0 \leq \alpha < 1$.

it is easy to use a barrier certificate $h$ to show the stability of the dynamical system. Let $\Ch = \{s: h(s) \geq 0\}$ be the superlevel set of $h$. The requirement of barrier certificates directly translates to the requirement that if $s \in \Ch$, then $f(s) \in \Ch$.
This property of $\Ch$, which is known as the \emph{forward-invariant} property, is especially useful in safety-critical settings: suppose a barrier certificate $h$ such that $\Ch$ does not contain unsafe states and contains the initial state $s_0$, then it is guaranteed that $\Ch$ contains the entire trajectory of states $\{s_t\}_{t \geq 0}$ which are safe.

Finding barrier certificates requires a known dynamics model $f$, which often can only be approximated in practice. This issue can be resolved by using a well-calibrated dynamics model $\hat f$, which predicts a confidence interval containing the true output. When a calibrated dynamics model $\hat f$ is used, we require that for any $s \in \States$, $\min_{s' \in \hat f(s)} h(s') \geq 0$.

Control barrier functions \citep{ames2019control} are extensions to barrier certificates in the control setting. That is, control barrier functions are often used to \emph{find} an action to meet the safety requirement instead of certifying the stability of a closed dynamical system. In this work, we simply use barrier certificates because in \Cref{sec:learn_barrier}, we view the policy and the calibrated dynamics model as a whole closed dynamical system whose stability we are going to certify.

\section{Learning Barrier Certificates via Adversarial Training}
\label{sec:learn_barrier}

This section describes an algorithm that learns a barrier certificate for a fixed policy $\pi$ under a calibrated dynamics model $\model$. Concretely, to certify a policy $\pi$ is safe, we aim to learn a (discrete-time) barrier certificate $\Lya$ that satisfies the following three requirements.

\begin{enumerate}[label=\textbf{R.\arabic*.},ref=\textbf{R.\arabic*},leftmargin=*,topsep=0pt,itemsep=2pt,partopsep=0pt, parsep=0pt]
    \item \label{itm:R1} For $s_0 \sim \mu$, $\Lya(s_0) \geq 0$ with probability 1.
    \item \label{itm:R2} For every $s \in \Sunsafe$, $\Lya(s) < 0$.
    \item \label{itm:R3} For any $s$ such that $\Lya(s) \geq 0$, $\min_{s' \in \model(s, \policy(s))}\Lya(s) \geq 0$.
\end{enumerate}
Requirement \ref{itm:R1} and \ref{itm:R3} guarantee that the policy $\policy$ will never leave the set  $\Ch = \{s \in \States: h(s) \geq 0\}$ by simple induction. Moreover, \ref{itm:R2} guarantees that $\Ch$ only contains safe states and therefore the policy never visits unsafe states.


In the rest of the section, we aim to design and train such a barrier certificate $\Lya=\Lya_\phi$ parametrized by neural network $\Lparam$.

\paragraph{$\Lya_\Lparam$ parametrization.} The three requirements for a \gLc are challenging to simultaneously enforce with constrained optimization involving neural network parameterization. Instead, we will parametrize $\Lya_\Lparam$ with \ref{itm:R1} and \ref{itm:R2} built-in such that for any $\Lparam$, $\Lya_\Lparam$ always satisfies \ref{itm:R1} and \ref{itm:R2}.

We assume the initial state $s_0$ is deterministic.
To capture the known user-specified safety set, we first handcraft a continuous function $\barrier: \States \to \RR_{\geq 0}$ satisfying $\barrier(s) \approx 0$ for typical $s \in \Ssafe$ and $\barrier(s) > 1$ for any $s \in \Sunsafe$ and can be seen as a smoothed indicator of $\Sunsafe$.\footnote{The function $\barrier(s)$ is called a barrier function for the user-specified safe set in the optimization literature. Here we do not use this term to avoid confusion with the barrier certificate. }
The construction of $\barrier$ does not need prior knowledge of irrecoverable states, but only the user-specified safety set $\Ssafe$.
To further encode the user-specified safety set into $\Lya_\Lparam$, we choose $\Lya_\Lparam$ to be of form $\Lya_\Lparam(s) = 1 - \text{Softplus}(f_\Lparam(s) - f_\Lparam(s_0)) - \barrier(s)$, where $f_\Lparam$ is a neural network, and $\text{Softplus}(x) = \log(1 + e^x)$.

Because $s_0$ is safe and $\barrier(s_0) \approx 0$, $\Lya_\Lparam(s_0) \approx 1 - \text{Softplus}(0) > 0$. Therefore $h_{\Lya}$ satisfies \ref{itm:R1}. Moreover, for any $s \in \Sunsafe$, we have $\Lya_\Lparam(s) < 1 - \barrier(s) < 0$, so $\Lya_\phi$ in our parametrization satisfies \ref{itm:R2} by design.

The parameterization can also be extended to multiple initial states. For example, if the initial $s_0$ is sampled from a distribution $\mu$ that is supported on a bounded set, and suppose that we are given the indicator function $\Binit: \States \to \RR$ for the support of $\mu$ (that is, $\Binit(s) = 1$ for any $s\in \textup{supp}(\mu)$, and $\Binit(s) = 0$ otherwise). Then, the parametrization of $h_\phi$ can be $h_\phi(s) = 1 - \text{Softplus}(f_\phi(s)) (1 - \Binit(s)) - \barrier(s)$. %
For simplicity, we focus on the case where there is a single initial state.

\paragraph{Training barrier certificates.} We now move on to training $\phi$ to satisfy \ref{itm:R3}.
Let
\begin{equation}
    \U(s, a, \Lya) := \max_{s' \in \model(s, a)} -\Lya(s').
    \label{eq:U}
\end{equation}
Then, \ref{itm:R3} requires $\U(s, \policy(s), \Lya_\Lparam) \leq 0$ for \textit{any} $s \in \levelset$,
The constraint in \ref{itm:R3} naturally leads up to formulate the problem as a min-max problem. Define our objective function to be
\begin{equation}
    C^*(\Lya_{\Lparam}, \U, \policy) := \max_{s \in \levelset} \U(s, \policy(s), \Lya_\Lparam) = \max_{s \in \levelset, s' \in \model(s, \pi(s))} -h(s')\,,
    \label{eq:C-def}
\end{equation}
and we want to minimize $C^*$ w.r.t. $\phi$:
\begin{align}
\min_{\Lparam} C^*(\Lya_\Lparam, \U, \policy) = \min_{\Lparam} \max_{s \in \levelset, s' \in \model(s, \pi(s))} -h(s'),
\end{align}
Our goal is to ensure the minimum value is less than 0.
We use gradient descent to solve the optimization problem, hence we need an explicit form of $\grad_\Lparam C^*$.
Let $L(s, \nu; \Lparam)$ be the Lagrangian for the constrained optimization problem in $C^*$ where $\nu \geq 0$ is the Lagrangian multiplier of the constraint $s \in \levelset$:
\[
    L(s, \nu; \Lparam) = U(s, \pi(s), h_\Lparam) + \nu h_\Lparam(s), \quad C^* = \max_{s \in \States} \min_{\nu \geq 0} L(s, \nu; \phi).
\]
By the Envelope Theorem (see Section 6.1 in \citet{carter2001foundations}), we have
\[
    \grad_\Lparam C^* = \grad_\Lparam U(s^*, \pi(s^*), h_\Lparam) + \nu^* \grad_\Lparam h_\Lparam(s^*),
\]
where $s^*$ and $\nu^*$ are the optimal solution for $\Lparam$. Once $s^*$ is known, the optimal Lagrangian multiplier $\nu^*$ can be given by KKT conditions:
\[
    \begin{cases}
        \nu^* h_\Lparam(s^*) = 0, \\
        \nabla_s L(s^*, \nu^*; \phi) = \textbf{0},
    \end{cases}
    \Longrightarrow \quad
    \nu^* = \begin{cases}
        0 & h_\phi(s^*) > 0, \\
        \frac{\|\grad_s \U(s^*, \policy(s^*), \Lya_\Lparam)\|_2}{\|\grad_s \Lya_\Lparam(s^*)\|_2} & h_\phi(s^*) = 0.
        \end{cases}
\]
Now we move on to the efficient calculation of $s^*$.

\paragraph{Computing the adversarial $s^*$.}
Because the maximization problem with respect to $s$ is nonconcave, there could be multiple local maxima.
In practice, we find that it is more efficient and reliable to use multiple local maxima to compute $\grad_\Lparam C^*$ and then average the gradient.

Solving $s^*$ is highly non-trivial, as it is a non-concave optimization problem with a constraint $s \in \levelset$.
To deal with the constraint, we introduce a Lagrangian multiplier $\lambda$ and optimize $U(s, \pi(s), \Lya_\Lparam) - \lambda \mathbb{I}_{s \in \levelset}$ w.r.t. $s$ without any constraints.
However, it is still very time-consuming to solve an optimization problem independently at each time.
Based on the observation that the parameters of $h$ do not change too much by one step of gradient step, we can use the optimal solution from the last optimization problem as the initial solution for the next one, which naturally leads to the idea of maintaining a set of candidates of $s^*$'s during the computation of $\grad_\Lparam C^*$.

We use Metropolis-adjusted Langevin algorithm (MALA) to maintain a set of candidates $\{s_{1}, \dots, s_m\}$ which are supposed to sample from $\exp(\tau (\U(s, \policy(s), \Lya_\Lparam) - \lambda \mathbb{I}_{s \in \levelset}))$.
Here $\tau$ is the temperature indicating we want to focus on the samples with large $\U(s, \policy(s), \Lya_\Lparam)$. Although the indicator function always have zero gradient, it is still useful in the sense that MALA will reject $s_i \not\in \levelset$. A detailed description of MALA is given in \Cref{sec:mala}.

We choose MALA over gradient descent because the maintained candidates are more diverse, approximate local maxima. If we use gradient descent to find $s^*$, then multiple runs of GD likely arrive at the same $s^*$, so that we lost the parallelism from simultaneously working with multiple local maxima. MALA avoids this issue by its intrinsic stochasticity, which can also be controlled by adjusting the hyperparameter $\tau$.

We summarize our algorithm of training barrier certificates in \Cref{alg:L-algo} (which contains optional regularization that will be discussed in \Cref{sec:regularization}). At Line~2, the initialization of $s_i$'s is arbitrary, as long as they have a sort of stochasticity.

\begin{algorithm}[tb]
    \begin{algorithmic}[1]
        \REQUIRE %
        Temperature $\tau$, Lagrangian multiplier $\lambda$, and optionally a regularization function $\mathsf{Reg}$.
		\STATE Let $U$ be defined as in Equation~\eqref{eq:U}.
        \STATE Initialize $m$ candidates of $s_{1}, \dots, s_{m}\in \States$ randomly.
        \FOR{$n$ iterations}
            \FOR{every candidate $s_i$}
                \STATE sample $s_i \sim \exp(\tau (U(s, \policy(s), \Lya_\Lparam) - \lambda \mathbb{I}_{s \in \Ch}))$ by MALA (\Cref{alg:mala}).
            \ENDFOR
                \STATE $W \gets \{s_i: \Lya_\Lparam(s_i) \geq 0, i \in [m]\}$.
                \STATE Train $\Lparam$ to minimize $C^*(\Lya_\Lparam, \U, \policy) + \mathsf{Reg}(\Lparam)$ using all candidates in $W$.
        \ENDFOR
    \end{algorithmic}
    \caption{Learning barrier certificate $\Lya_\Lparam$ for a policy $\policy$ w.r.t. a calibrated dynamics model $\model$. }
    \label{alg:L-algo}
\end{algorithm}

\section{CRABS: \textbf{C}o-t\textbf{ra}ined \textbf{B}arrier Certificate for \textbf{S}afe RL}
\label{sec:crabs}

In this section, we present our main algorithm, \textbf{C}o-t\textbf{ra}ined \textbf{B}arrier Certificate for \textbf{S}afe RL (CRABS), shown in \Cref{alg:paradigm}, to \emph{iteratively} co-train barrier certificates, policy and dynamics model, using the algorithm in \Cref{sec:learn_barrier}. In addition to parametrizing $\Lya$ by $\Lparam$, we further parametrize the policy $\policy$ by $\Piparam$, and parametrize calibrated dynamics model $\model$ by $\Mparam$.
CRABS alternates between training a \gLc that certifies the policy $\policy_\Piparam$ w.r.t. a calibrated dynamics model $\model_\Mparam$ (Line~5), collecting data safely using the certified policy (Line~3, details in \Cref{sec:impl-data-collection}), learning a calibrated dynamics model (Line~4, details in \Cref{sec:impl-model}), and training a policy with the constraint of staying in the superlevel set of the barrier function (Line~6, details in \Cref{sec:impl-policy-opt}).%
In the following subsections, we discuss how we implement each line in detail.

\begin{algorithm}[tb]
    \caption{CRABS: \textbf{C}o-t\textbf{ra}ined \textbf{B}arrier Certificate for \textbf{S}afe RL (Details in \Cref{sec:crabs})}
	\begin{algorithmic}[1]
		\REQUIRE An initial safe policy $\piinit$.
		\STATE Collected trajectories buffer $\buffer\gets \emptyset$; $\policy \gets \piinit$.
		\FOR{$T$ epochs}
		\STATE \label{line:data-collection} Invoke \Cref{alg:safe-exploration} to safely collect trajectories (using $\pi$ as the safeguard policy and a noisy version of $\pi$ as the $\piexpl$).  Add the trajectories to $\buffer$.
		\STATE \label{line:learn-dynamics} Learn a calibrated dynamics model $\model$ with $\buffer$.
		\STATE \label{line:learn-L} Learn a \gLc $\Lya$ that certifies $\policy$ w.r.t. $\model$ using \Cref{alg:L-algo} with regularization.
		\STATE \label{line:2} Optimize policy $\pi$ (according to the reward), using data in $\buffer$, with the constraint that $\pi$ is certified by $\Lya$.
		\ENDFOR
	\end{algorithmic}

	\label{alg:paradigm}
\end{algorithm}

\subsection{Safe Exploration with Certified Safeguard Policy}
\label{sec:impl-data-collection}

\begin{algorithm}[t]
    \begin{algorithmic}[1]
        \REQUIRE (1) A policy $\backuppolicy$ certified by \gLc $\Lya$, (2) any proposal exploration policy $\piexpl$.
        \REQUIRE A state $s \in \levelset$.
            \STATE Sample $n$ actions $a_{1}, \dots a_{n}$ from $\piexpl(s)$.
            \IF{there exists an $a_i$ such that $\U(s, a_i, \Lya) \leq 1$}
            \STATE    \textbf{return:} $a_i$ %
            \ELSE
                \STATE \textbf{return:} $\backuppolicy(s)$.
            \ENDIF
    \end{algorithmic}
    \caption{Safe exploration with safeguard policy $\backuppolicy$}
    \label{alg:safe-exploration}
\end{algorithm}

Safe exploration is challenging because it is difficult to detect irrecoverable states.
The barrier certificate is designed to address this---a policy $\pi$ certified by some $\Lya$ guarantees to stay within $\Ch$ and therefore can be used for collecting data. %
However, we may need more diversity in the collected data beyond what can be offered by the deterministic certified policy $\backuppolicy$. Thanks to the contraction property~\ref{itm:R3}, we in fact know that any exploration policy $\piexpl$ within the superlevel set $\Ch$ can be made safe with $\backuppolicy$ being a safeguard policy---we can first try actions from $\piexpl$ and see if they stay within the viable subset $\Ch$, and if none does, invoke the safeguard  policy $\backuppolicy$.
\Cref{alg:safe-exploration} describes formally this simple procedure that makes any exploration policy $\piexpl$ safe. By a simple induction, one can see that the policy defined in \Cref{alg:safe-exploration} maintains that all the visited states lie in $\Ch$. The main idea of \Cref{alg:safe-exploration} is also widely used in policy shielding \citep{alshiekh2018safe,jansen2018safe,anderson2020neurosymbolic}, as the policy $\backuppolicy$ sheilds the policy $\pi$ in $\levelset$.

The safeguard policy $\backuppolicy$ is supposed to safeguard the exploration.
However, activating the safeguard too often is undesirable, as it only collects data from $\backuppolicy$ so there will be little exploration.
To mitigate this issue, we often choose $\piexpl$ to be a noisy version of $\backuppolicy$ so that $\piexpl$ will be roughly safe by itself. Moreover, the safeguard policy $\backuppolicy$ will be trained via optimizing the reward function as shown in the next subsections.
Therefore, a noisy version of $\backuppolicy$ will explore the high-reward region and avoid unnecessary exploration.

Following \citet{haarnoja2018soft}, the policy $\policy_\Piparam$ is parametrized as $\tanh(\mu_\Piparam(s))$, and the proposal exploration policy $\piexpl_\Piparam$ is parametrized as $\tanh(\mu_\Piparam(s) + \sigma_\Piparam(s) \zeta)$ for $\zeta \sim \mathcal{N}(0, I)$, where $\mu_\Piparam$ and $\sigma_\Piparam$ are two neural networks.
Here the $\tanh$ is applied to squash the outputs to the action set $[-1,1]$.

\subsection{Regularizing Barrrier Certificates}
\label{sec:regularization}
The quality of exploration is directly related to the quality of policy optimization.
In our case, the exploration is only within the learned viable set $\levelset$ and it will be hindered if $\levelset$ is too small or does not grow during training. To ensure a large and growing viable subset $\levelset$, we encourage the volume of $\levelset$ to be large by adding a regularization term%
$$
\mathsf{Reg}(\Lparam; \hat\Lya) = \mathbb{E}_{s \in \States} [\textup{relu}(\hat\Lya(s) - \Lya_\Lparam(s))],
$$
Here $\hat\Lya$ is the barrier certificate obtained in the previous epoch. In the ideal case when $\mathsf{Reg}(\Lparam; \hat\Lya) = 0$, we have $ \level_{\Lya_\Lparam}  \supset \level_{\hat\Lya}$, that is, the new viable subset $\level_{\Lya_\Lparam} $ is at least bigger than the reference set (which is the viable subset in the previous epoch.)   %
We compute the expectation over $\States$ approximately by using the set of candidate $s$'s maintained by MALA.  %

In summary, to learn $\Lya_{\Lparam}$ in CRABS, we minimize the following objective (for a small positive constant $\lambda$) over $\Lparam$ as shown in \Cref{alg:L-algo}:
\begin{equation}
    \mathcal{L}(\Lparam; \U, \policy_\Piparam, \hat\Lya) = C^*(L_\Lparam, \U, \policy_\Piparam) + \lambda \mathsf{Reg}(\Lparam; \hat\Lya).
    \label{eq:L-loss}
\end{equation}

We remark that the regularization is not the only reason why the viable set $\levelset$ can grow. When the dynamics model becomes more accurate as we collect more data, the $\levelset$ will also grow.
This is because an inaccurate dynamics model will typically make the $\levelset$ smaller---it is harder to satisfy \ref{itm:R3} when the confidence region $\model(s, \pi(s))$ in the constraint contains many possible states. Vice versa, shrinking the size of the confidence region will make it easier to certify more states.

\subsection{Learning a Calibrated Dynamics Model}
\label{sec:impl-model}
It is a challenging open question to obtain a dynamics model $\model$ (or any supervised learning model) that is theoretically well-calibrated especially with domain shift~\cite{zhao2020individual}.
In practice, we heuristically approximate a calibrated dynamics model by learning an ensemble of probabilistic dynamics models, following common practice in RL~\citep{yu2020mopo,janner2019trust,chua2018deep}. We learn $K$ probabilistic dynamics models $f_{\Mparam_{1}}, \dots, f_{\Mparam_{K}}$ using the data in the replay buffer $\buffer$.
(Interestingly, prior work shows that an ensemble of probabilistic models can still capture the error of estimating a deterministic ground-truth dynamics model \citep{janner2019trust,chua2018deep}.)
Each probabilistic dynamics model $f_{\Mparam_i}$ outputs a Gaussian distribution $\mathcal{N}(\mu_{\Mparam_i}(s, a), \diag(\sigma^2_{\Mparam_i}(s, a)))$ with diagonal covariances, where $\mu_{\Mparam_i}$ and $\sigma_{\Mparam_i}$ are parameterized by neural networks.
Given a replay buffer $\buffer$, the objective for a probabilistic dynamics model $f_{\Mparam_i}$ is to minimize the negative log-likelihood:
\begin{equation}
    \mathcal{L}_{\model} (\Mparam_i) = -\mathbb{E}_{(s, a, s') \sim \buffer} \left[-\log f_{\Mparam_i}(s'|s, a) \right].
    \label{eq:model-loss}
\end{equation}
The only difference in the training procedure of these probabilistic models is the randomness in the initialization and mini-batches. We simply aggregate the means of all learn dynamics models as a coarse approximation of the confidence region, i.e., $\model(s, a) = \{\mu_{\Mparam_i}(s, a)\}_{i \in [K]}$.

We note that we implicitly rely on the neural networks for the dynamics model to extrapolate to unseen states. However, local extrapolation suffices. The dynamics models' accuracy affects the size of the viable set---the more accurate the model is, the more likely the viable set is bigger. In each epoch, we rely on the additional data collected and the model's extrapolation to reduce the errors of the learned dynamics model on unseen states that are \textit{near} the seen states, so that the learned viable set can grow in the next epoch. Indeed, in \Cref{sec:experiment} (\Cref{fig:recoverable-states}) we show that the viable set grows gradually as the error and uncertainty of the models improves over epoch.

\subsection{Policy Optimization}
\label{sec:impl-policy-opt}

We describe our policy optimization algorithm in \Cref{alg:safe-SAC}. The desiderata here are (1) the policy needs certified by the current \gLc $\Lya$ and (2) the policy has as high reward as possible. We break down our policy optimization algorithm into two components: First, we optimize the total rewards $\return(\policy_\Piparam)$ of the policy $\policy_\Piparam$; Second, we use adversarial training to guarantee the optimized policy can be certified by $\Lya_\Lparam$.
The modification of SAC is to some extent non-essential and mostly for technical convenience of making SAC somewhat compatible with the constraint set. Instead, it is the adversarial step that fundamentally guarantees that the policy is certified by the current $\Lya_\Lparam$.

\paragraph{Adversarial training.}

We use adversarial training to guarantee $\policy_\Piparam$ can be certified by $\Lya_\Lparam$.
Similar to what we've done in training $\Lya_\Lparam$ adversarially, the objective for training $\policy_\Piparam$  is to minimize $C^*(\Lya_\Lparam, \U, \policy_\Piparam)$.
Unlike the case of $\Lparam$, the gradient of $C^*(\Lya_\Lparam, \U, \policy_\Piparam)$ w.r.t. $\Piparam$ is simply $\grad_\Piparam \U(s^*, \policy_\Piparam(s^*), \Lya_\Lparam)$, as the constraint $\Lya_\Lparam(s)$ is unrelated to $\policy_\Piparam$.
We also use MALA to solve $s^*$ and plug it into the gradient term $\grad_\Piparam \U(s^*, \policy_\Piparam(s^*), \Lya_\Lparam)$.

\paragraph{Optimizing $\return(\policy_\Piparam)$.}
We use a modified SAC \citep{haarnoja2018soft} to optimize $\return(\policy_\Piparam)$. As the modification is for safety concerns and is minor, we defer it to \Cref{sec:reward-opt}. As a side note, although we only optimize $\piexpl_\Piparam$ here, $\policy_\Piparam$ is also optimized implicitly because $\piexpl_\Piparam$ simply outputs the mean of $\pi_\Piparam$ deterministically.

\section{High-risk, High-reward Environments}

\begin{figure}[t]
    \begin{minipage}[c]{0.58\textwidth}
        \centering
        \begin{subfigure}[b]{0.3\linewidth}
            \centering
            \includegraphics[width=\linewidth]{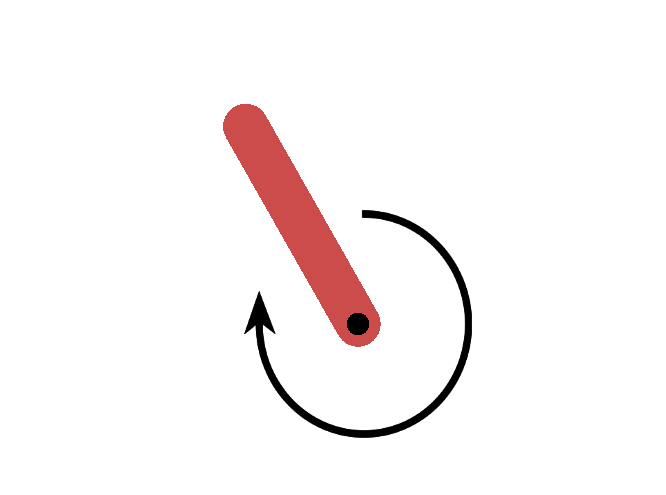}
            \caption{Pendulum}
            \label{fig:invpen}
        \end{subfigure}
        \begin{subfigure}[b]{0.3\linewidth}
            \centering
            \includegraphics[width=\linewidth]{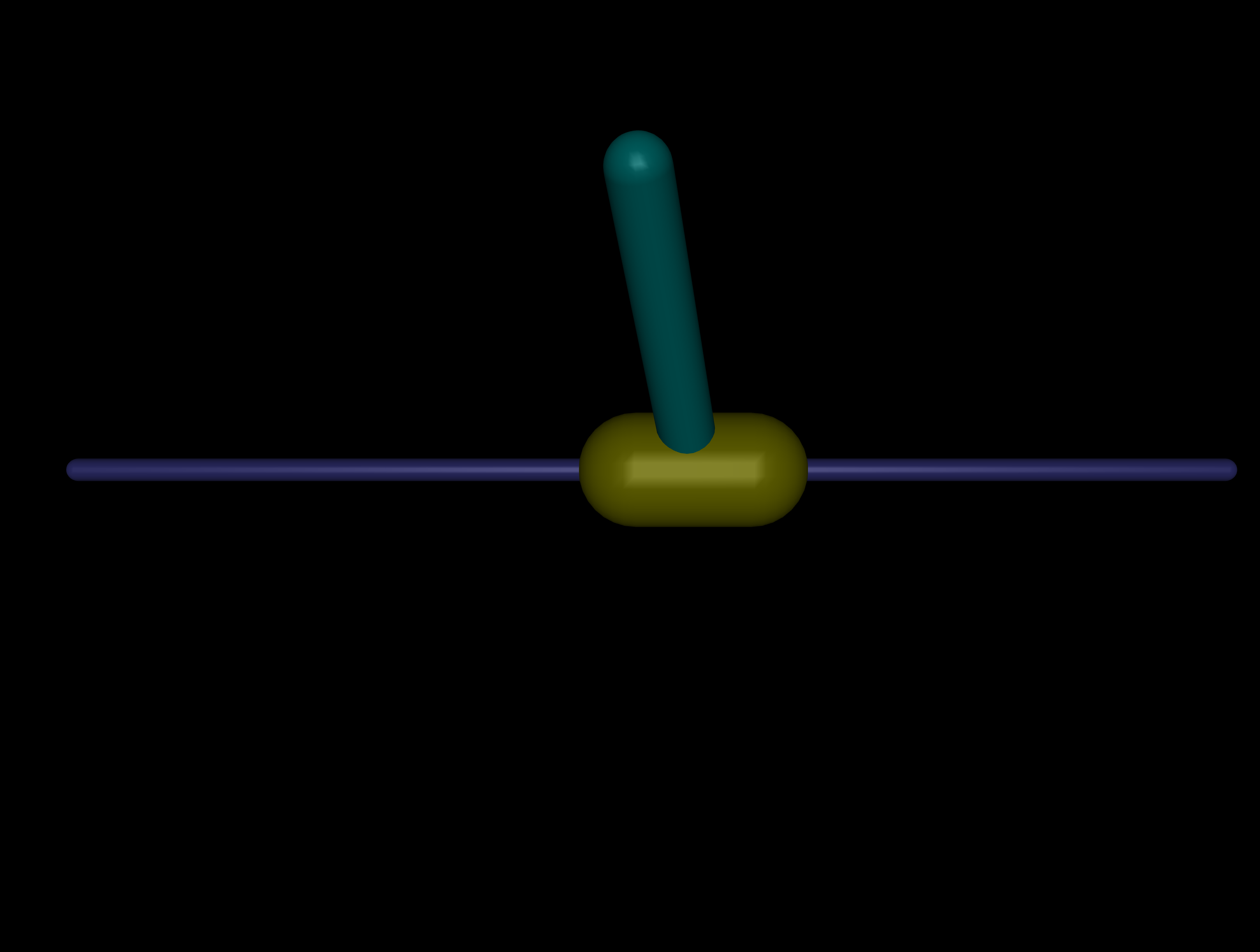}
            \caption{CartPole}
            \label{fig:cartpole}
        \end{subfigure}
    \end{minipage}\hfill
    \begin{minipage}[c]{0.4\textwidth}
        \caption{Illustration of environments. The left figure illustrates the Pendulum environment, which is used by \env{Upright} and \env{Tilt} tasks. The right figer illustrates the CartPole environment, which is used by \env{Move} and \env{Swing} tasks. %
        }
        \label{fig:illustration-env}
    \end{minipage}
  \end{figure}

We design four tasks, three of which are high-risk, high-reward tasks, to check the efficacy of our algorithm. Even though they are all based on inverted pendulum or cart pole, we choose the reward function to be somewhat conflicted with the safety constraints. That is, the optimal policy needs to take a trajectory that is near the safety boundary. This makes the tasks particularly challenging and suitable for stress testing our algorithm's capability of avoiding irrecoverable states.

These tasks have state dimension dimensions between 2 to 4. We focus on the relatively low dimensional environments to avoid conflating the failure to learn accurate dynamics models from data and the failure to provide safety given a learned approximate dynamics model.
Indeed, we identify that the major difficulty to scale up to high-dimensional environments is that it requires significantly more data to learn a decent high-dimensional dynamics model that can predict long-horizon trajectories.
We remark that we aim to have zero violations. This is very difficult to achieve, even if the environment is low dimensional. As shown by \Cref{sec:experiment}, many existing algorithms fail to do so.

(a) \env{Upright}. The task is based on Pendulum-v0 in Open AI Gym~\cite{gym}, as shown in \Cref{fig:invpen}.
The agent can apply torque to control a pole. The environment involves the crucial quantity: the tilt angle $\theta$ which is defined to be the angle between the pole and a vertical line. The safety requirement is that the pole does not fall below the horizontal line. Technically, the user-specified safety set is  $\{\theta: |\theta|\le \theta_\text{max} = 1.5\}$ (note that  the threshold is very close to $\frac{\pi}{2}$ which corresponds to 90$^\circ$.)
The reward function $r$ is $r(s, a) = -\theta^2$, so the optimal policy minimizes the angle and angular speed by  keeping the pole upright. The horizon is 200 and the initial state $s_0 = (0.3, -0.9)$.

(b) \env{Tilt}. This action set, dynamics model, and horizon, and safety set are the same as in \textit{Upright}. The reward function is different: $r(s, a) = -(\theta_\text{limit} - \theta)^2$.
The optimal policy is supposed to stay tilting near the angle $\theta=\theta_\text{limit}$ where $\theta_\text{limit} = -0.41151684$ is the largest angle the pendulum can stay balanced. The challenge is during exploration, it is easy for the pole to overshoot and violate the safety constraints.

(c) \env{Move}. The task is based on a cart pole and the goal is to move a cart (the yellow block) to control the pole (with color teal), as shown in \Cref{fig:cartpole}.
The cart has an $x$ position between $-1$ and $1$, and the pole also has an angle $\theta \in [-\frac{\pi}{2}, \frac{\pi}{2}]$ with the same meaning as \env{Upright} and \env{Tilt}.
The starting position is $x= \theta = 0$.
We design the reward function to be $r(s, a) = x^2$. The user-specified safety set is  $\{(x, \theta): |\theta|\le \theta_\text{max} = 0.2, |x| \leq 0.9 \}$ where 0.2 corresponds to roughly 11$^\circ$.
Therefore, the optimal policy needs to move the cart and the pole slowly in one direction, preventing the pole from falling down and the cart from going too far.
The horizon is set to 1000.

(d) \env{Swing}.
This task is similar to \env{Move}, except for a few differences: The reward function is $r(s, a) = \theta^2$; The user-specified safety set is $\{(x, \theta): |\theta| \leq \theta_\text{max} = 1.5, |x| \leq 0.9\}$. So the optimal policy will swing back and forth to some degree and needs to control the angles well so that it does not violate the safety requirement. %

For all the tasks, once the safety constraint is violated, the episode will terminate immediately and the agent will receive a reward of -30 as a penalty. The number -30 is tuned by running SAC and choosing the one that SAC performs best with.

\section{Experimental  Results}
\label{sec:experiment}

\begin{figure}[t]
    \centering
    \includegraphics[width=\textwidth]{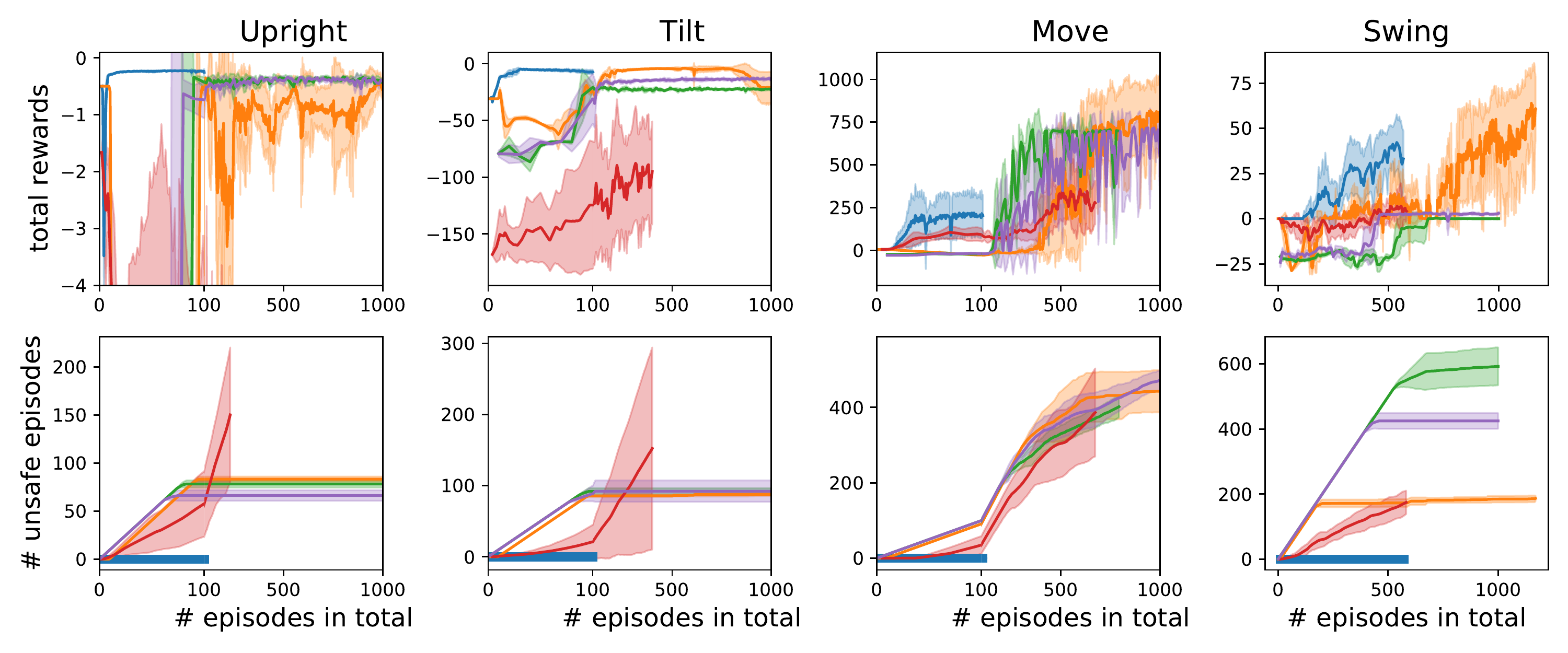} \\
    \includegraphics[width=0.9 \textwidth]{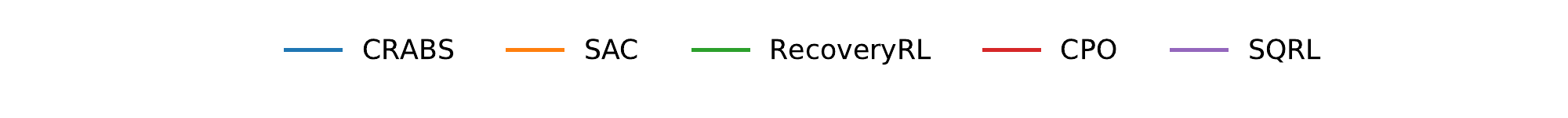}
    \captionof{figure}{
        Comparision between \algo{} and baselines. \algo{} can learn a policy without any safety violations, while other baselines have a lot of safety violations.
    We run each algorithm four times with independent randomness. The solid curves indicate the mean of four runs and the shaded areas indicate one standard deviation around the mean.
    }
    \label{fig:main-results}
\end{figure}

\begin{figure}[t]
    \centering
    \includegraphics[width=\textwidth]{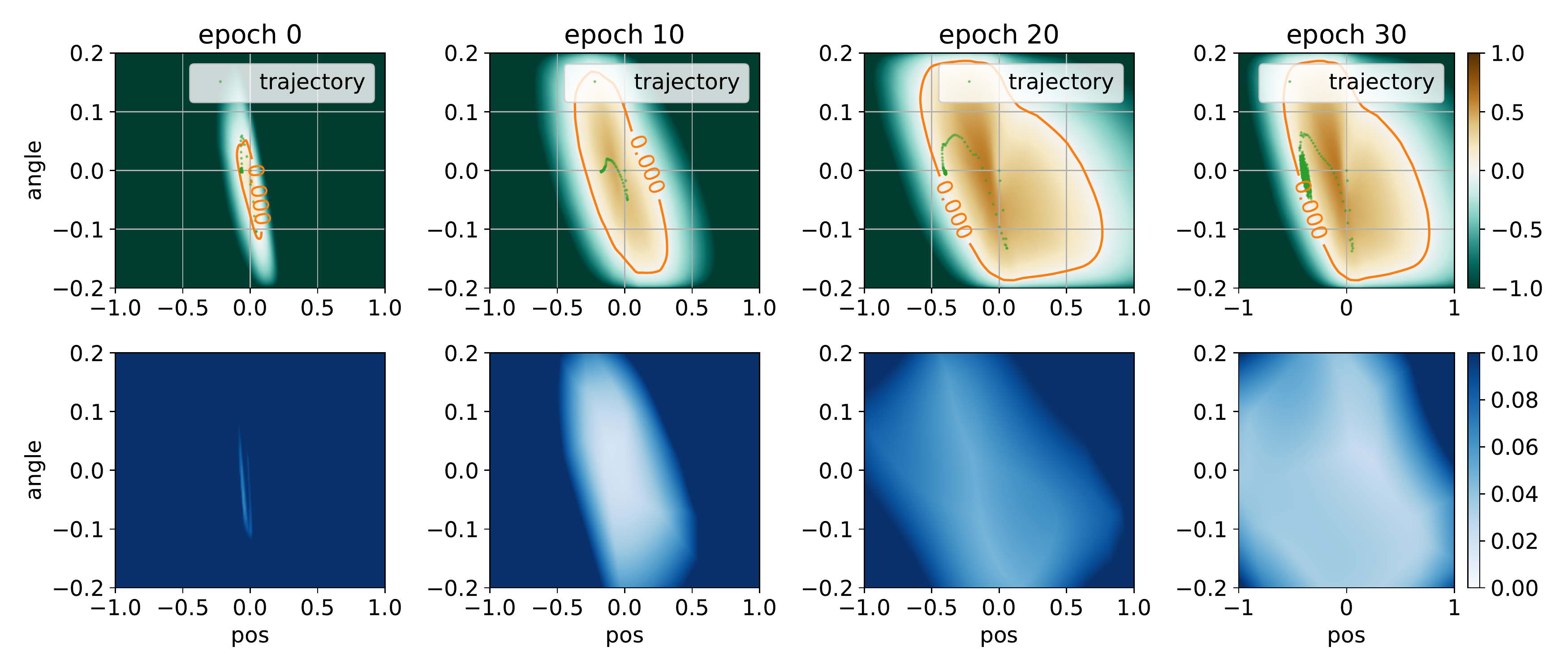}
    \caption{Visualization of the growing viable subsets learned by \algo{} in \env{Move}. To illustrate the 4-dimensional state space, we project a state from $[x, \theta, \dot x, \dot \theta]$ to $[x, \theta]$.
    \textbf{Top} row illustrates the learned $\levelset$ at different epochs. The red curve encloses superlevel set $\levelset$, while the green points indicate the projected trajectory of the current safe policy. We can also observe that policy $\policy$ learns to move left as required by the task.
   We note that shown states in the trajectory sometimes seemingly are not be enclosed by the red curve due to the projection. %
   \textbf{Bottom} row illustrates the size of confidence region (defined by \Cref{eq:uncertainty}) of the dynamics model $\model_{\Mparam}$ at different epochs. Darker states mean the learned dynamics model is less confident. We can observe that the learned dynamics model tends to be confident at more states after diverse data are collected.
   }
    \label{fig:recoverable-states}
\end{figure}

In this section, we conduct experiments to answer the following question: Can CRABS learn a reasonable policy without safety violations in the designed tasks?

\paragraph{Baselines.}
We compare our algorithm \algo{} against four baselines:
(a) \textbf{Soft Actor-Critic} (SAC) \citep{haarnoja2018soft}, one of the state-of-the-art RL algorithms,
(b) \textbf{Constrained Policy Optimization} (CPO) \citep{achiam2017constrained}, a safe RL algorithm which builds a trust-region around the current policy and optimizes the policy in the trust-region,
(c) \textbf{RecoveryRL} \citep{thananjeyan2021recovery} which leverages offline data to pretrain a risk-sensitive $Q$ function and also utilize two policies to achieving two goals (being safe and obtaining high rewards),
and (d) \textbf{SQRL} \citep{srinivasan2020learning} which leverages offline data in an easier environment and fine-tunes the policy in a more difficult environment.
SAC and CPO are given an initial safe policy for safe exploration, while RecoveryRL and SQRL are given offline data containing 40K steps from both mixed safe and unsafe trajectories which are free and are not counted. CRABS collects more data at each iteration in \env{Swing} than in other tasks to learn a better dynamics model $\model$.
For SAC, we use the default hyperparameters because we found they are not sensitive. For RecoveryRL and SQRL, the hyperparameters are tuned in the same way as in \citet{thananjeyan2021recovery}. For CPO, we tune the step size and batch size.
More details of experiment setup and the implementation of baselines can be found in \Cref{sec:exp-details}.

\paragraph{Results.}
Our main results are shown in \Cref{fig:main-results}.
From the perspective of total rewards, SAC achieves the best total rewards among all of the 5 algorithms in \env{Move} and \env{Swing}.
 In all tasks, CRABS can achieve reasonable total rewards and learns faster at the beginning of training, and we hypothesize that this is directly due to its strong safety enforcement. RecoveryRL and SQRL learn faster than SAC in \env{Move}, but they suffer in \env{Swing}. RecoveryRL and SQRL are not capable of learning in \env{Swing}, although we observed the average return during exploration at the late stages of training can be as high as 15. %
CPO is quite sample-inefficient and does not achieve reasonable total rewards as well.

From the perspective of safety violations, CRABS surpasses all baselines \textbf{without a single safety violation}.
The baseline algorithms always suffer from many safety violations. SAC, SQRL, and RecoveryRL have a similar number of unsafe trajectories in \env{Upright, Tilt, Move}, while in \env{Swing}, SAC has the fewest violations and RecoveryRL has the most violations.
CPO has a lot of safety violations. We observe that for some random seeds, CPO does find a safe policy and once the policy is trained well, the safety violations become much less frequent, but for other random seeds, CPO keeps visiting unsafe trajectories before it reaches its computation budget.

\paragraph{Visualization of learned viable subset $\levelset$.} To demonstrate that the algorithms work as expected, we visualized the viable set $\levelset$ in \Cref{fig:recoverable-states}.
As shown in the figure, our algorithm \algo{} succeeds in certifying more and more viable states and does not get stuck locally, which demonstrates the efficacy of the regularization at \Cref{sec:regularization}. We also visualized how confident the learned dynamics model is as training goes on. More specifically, the uncertainty of a calibrated dynamics model $\model$ at state $s$ is defined as
\begin{equation}
    \textsf{Uncertainty}(\model, s) = \max_{s_1, s_2 \in \model(s, \textbf{0})} \|s_1 - s_2\|_2.
    \label{eq:uncertainty}
\end{equation}
We can see from \Cref{fig:recoverable-states} that the initial dynamics model is only locally confident around the initial policy, but becomes more and more confident after collecting more data.

\paragraph{Handcrafted barrier function $h$.} To demonstrate the advantage of learning a barrier function, we also conduct experiments on a variant of CRABS, which uses a handcrafted barrier certificate by ourselves and does not train it, that is, \Cref{alg:paradigm} without Line 5. The results show that this variant does not perform well: It does not achieve high rewards, and has many safety violations. We hypothesize that the policy optimization is often burdened by adversarial training, and the safeguard policy sometimes cannot find an action to stay within the superlevel set $\Ch$.

\section{Related Work}

Prior works about Safe RL take very different approaches. \citet{dalal2018safe} adds an additional layer, which corrects the output of the policy locally.
Some of them use Lagrangian methods to solve CMDP, while the Lagrangian multiplier is controlled adaptively \citep{tessler2018reward} or by a PID \citep{stooke2020responsive}. \citet{achiam2017constrained,yang2020accelerating} build a trust-region around the current policy, and \citet{zanger2021safe} further improved \citet{achiam2017constrained} by learning the dynamics model.
\citet{eysenbach2017leave} learns a reset policy so that the policy only explores the states that can go back to the initial state.
\citet{turchetta2020safe} introduces a learnable teacher, which keeps the student safe and helps the student learn faster in a curriculum manner.
\citet{srinivasan2020learning} pre-trains a policy in a simpler environment and fine-tunes it in a more difficult environment.
\citet{bharadhwaj2020conservative} learns conservative safety critics which underestimate how safe the policy is, and uses the conservative safety critics for safe exploration and policy optimization.
\citet{thananjeyan2021recovery} makes use of existing offline data and co-trains a recovery policy.

Another line of work involves Lyapunov functions and barrier functions. %
\citet{chow2018lyapunov} studies the properties of Lyapunov functions and learns them via bootstrapping with a discrete action space. Built upon \citet{chow2018lyapunov}, \citet{sikchi2021lyapunov} learns the policy with Deterministic Policy Gradient theorem in environments with a continuous action space. Like TRPO \citep{schulman2015trust}, \citet{sikchi2021lyapunov} also builds a trust region of policies for optimization.
\citet{donti2020enforcing} constructs sets of stabilizing actions using a Lyapunov function, and project the action to the set, while \citet{chow2019lyapunov} projects action or parameters to ensure the decrease of Lyapunov function after a step.
\citet{ohnishi2019barrier} is similar to ours but it constructs a barrier function manually instead of learning such one.
\citet{ames2019control} gives an excellent overview of control barrier functions and how to design them. Perhaps the most related work to ours is \citet{cheng2019end}, which also uses a barrier function to safeguard exploration and uses a reinforcement learning algorithm to learn a policy. However, the key difference is that we \emph{learn} a barrier function, while \citet{cheng2019end} handcrafts one. The works on Lyapunov functions~\cite{berkenkamp2017safe,richards2018lyapunov} require the discretizating the state space and thus only work for low-dimensional space.

\citet{anderson2020neurosymbolic} iteratively learns a neural policy which possily has higher total rewards but is more unasfe, distills the learned neural policy into a symbolic policy which is simpler and safer, and use automatic verification to certify the symbolic policy. The certification process is similar to construct a barrier function. As the certification is done on a learned policy, the region of certified states also grows. Howver, it assumes a known calibrated dynamcis model, while we also learns it. Also, tt can only certifie states where a piecewise-linear policy is safe, while potentially we can certify more states.

\section{Conclusion}

In this paper, we propose a novel algorithm \algo{} for training-time safe RL. The key idea is that we co-train a barrier certificate together with the policy to certify viable states, and only explore in the learned viable subset.
The empirical rseults show that \algo{} can learn some tasks without a single safety violation. We consider using model-based policy optimization techniques to improve the total rewards and sample efficiency as a promising future work.

We focus on low-dimensional continuous state space in this paper because it is already a sufficiently challenging setting for zero training-time violations, and we leave the high-dimensional state space as an important open question. We observed in our experiments that it becomes more challenging to learn a dynamics model in higher dimensional state space that is sufficiently  accurate and calibrated even under the training data distribution (the distribution of observed trajectories). Therefore, to extend our algorithms to high dimensional state space, we suspect that we either need to learn better dynamics models or the algorithm needs to be more robust to the errors in the dynamics model.

\section*{Acknowledgement}

We thank Changliu Liu, Kefan Dong, and Garrett Thomas for their insightful comments. YL is supported by NSF, ONR, Simons Foundation, Schmidt Foundation, Amazon Research, DARPA and SRC. TM acknowledges support of Google Faculty Award,  NSF IIS 2045685, and JD.com.

\bibliography{ref,refs/all.bib}
\bibliographystyle{plainnat}

\appendix

\newpage
\section{Reward Optimizing in CRABS}
\label{sec:reward-opt}

As in original SAC, we maintain two $Q$ functions $Q_{\Qparam_i}$and their target networks $Q_{\bar \Qparam_i}$ for $i \in \{1, 2\}$, together with a learnable temperature $\alpha$. The objective for the policy is to minimize
\begin{equation}
    \begin{aligned}
        \mathcal{L}_\pi(\Piparam) &= \mathbb{E}_{s \sim \buffer, a \sim \policy_\Piparam} \left[ \alpha \log \piexpl_{\Piparam}(a | s) - \hat Q_{\Qparam_1}(s, a) \right],
    \end{aligned}
    \label{eq:pi-loss}
\end{equation}
where $\hat Q_{\Qparam_1}(s, a) = Q_{\Qparam_1}(s, a)$ if $\U(s, a, \Lya) \leq 0$, otherwise $\hat Q_{\Qparam_1}(s, a) = -C - \U(s, a, \Lya)$ for a large enough constant $C$. The heuristics behind the design of $\hat Q_{\Qparam_1}$ is that we should lower the probability of $\piexpl_\Piparam$ proposing an action which will possibly leave the superlevel set $\levelset$ to reduce the frequency of invoking the safeguard policy during exploration.

The temporal difference objective for the $Q$ function is
\begin{equation}
        \mathcal{L}_Q(\Qparam_i) = \mathbb{E}_{(s, a, r, s') \sim \buffer} \mathbb{E}_{a' \sim \piexpl_\Piparam(s')} \left[(Q_{\Qparam_i}(s, a) - (r + \gamma \min_{i \in \{1, 2\}}Q_{\bar\Qparam_i}(s, a)))^2 \mathbb{I}_{\U(s', a', \Lya_\Lparam) \leq 0} \right],
    \label{eq:Q-loss}
\end{equation}
We remark that we reject all $a' \sim \piexpl_\Piparam(s')$ such that $\U(s', a', \Lya_\Lparam) > 0$, as our safe exploration algorithm (\Cref{alg:safe-exploration}) will reject all of them eventually.
The temperature $\alpha$ is learned the same as in \citet{haarnoja2018soft}:
\begin{equation}
   \mathcal{L}_\alpha(\alpha) = \mathbb{E}_{s \sim \buffer} [-\alpha \log \piexpl_\Piparam(a | s) - \alpha \bar{\mathcal{H}}],
   \label{eq:alpha-loss}
\end{equation}
where $\bar{\mathcal{H}}$ is hyperparameter, indicating the target entropy of the policy $\piexpl_\Piparam$.

\begin{algorithm}[tb]
    \begin{algorithmic}[1]
        \INPUT A policy $\pi$, the replay buffer $\buffer$
        \STATE Sample a batch $\batch$ from buffer $\buffer$.
        \STATE Train $\Piparam$ to minimize $\mathcal{L}_\policy(\Piparam)$ using $\batch$.
        \STATE Train $Q$ to minimize $\mathcal{L}_Q(\Qparam_i)$ for $i \in \{1, 2\}$ using $\batch$.
        \STATE Train $\alpha$ to minimize $\mathcal{L}_\alpha(\alpha)$ using $\batch$.
        \STATE Invoke MALA to training $s^*$ adversarially (as in L4-5 in \Cref{alg:L-algo}).
        \STATE Train $\Piparam$ minimize $C^*(\Lya_\Lparam, \U, \policy_\Piparam)$.
        \STATE Update target network $\bar\Qparam_{i}$ for $i \in \{1, 2\}$.
    \end{algorithmic}
    \caption{Modified SAC to train a policy while constraining it to stay within $\levelset$}
    \label{alg:safe-SAC}
\end{algorithm}

\section{Experiment Details}
\label{sec:exp-details}

Our code is implemented by Pytorch \citep{NEURIPS2019_9015} and runs in a single RTX-2080 GPU. Typically it takes 12 hours to run one seed for \env{Upright}, \env{Tilt} and \env{Move}, and for \env{Swing} it takes around 60 hours. In a typical run of \env{Swing}, 33 hours are spent on learning barrier functions.

\subsection{Environment}
\label{sec:env-details}

All the environments are based on OpenAI Gym \citep{gym} where MuJoCo \citep{todorov2012mujoco} serves as the underlying physics engine. We use discount $\gamma = 0.99$.

The tasks \env{Upright} and \env{Tilt} are based on \verb|Pendulum-v0|. The obsevation is $[\theta, \dot\theta]$ where $\theta$ is the angle between the pole and a vertical line, and $\dot\theta$ is the angular velocity. The agent can apply a torque to the pendulum.
The task \env{Move} and \env{Swing} is based on \verb|InvertedPendulum-v2| with observation $[x, \theta, \dot x, \dot \theta]$.
The agent can control how the cart moves.

As all of the constraints are in the form of $\|\theta\| \leq \theta_\text{max}$ and $|x| \leq x_\text{max}$. For each type of constraint, we design $\barrier$ to be
\[
    \barrier(s) = \max\left( \omega\left(\theta/\theta_\text{max} \right), \omega\left( x / x_\text{max} \right) \right),
\]
with $\omega(x) = \max(0, 100(|x| - 1))$. If there is no constraint of $x$, we just take $\barrier(s) = \omega\left(\theta / \theta_\text{max} \right)$.
One can easy check that $\barrier(s)$ is continuous and equals to 1 at the boundary of safety set.

\subsection{Hyperparameters}

\paragraph{Policy}
We parametrize our policy using a feed-forward neural network with ReLU activation and two hidden layers, each of which contains 256 hidden units. Similar to \citet{haarnoja2018soft}, the output of the policy is squashed by a tanh function.

The initial policy is obtained by running SAC for $10^5$ steps, checking the intermediate policy for every $10^4$ steps and picking the first safe intermediate policy.

In all tasks, we optimize the policy for 2000 steps in a single epoch.

\paragraph{Dynamics Model}
We use an ensemble of five learned dynamics models as the calibrated dynamcis model. Each of the dynamics model contains 4 hidden layers with 400 hidden units and use Swish as the activation function \citep{ramachandran2017searching}. Following \citet{chua2018deep}, we also train learnable parameters to bound the output of $\sigma_\Mparam$. We use Adam \citep{kingma2014adam} with learning rate 0.001, weight decay 0.000075 and batch size 256 to optimize the dynamics model.

In the experiment \env{Move} and \env{Swing}, the initial model is obtained by traininng one a data for 20000 steps with 500 safe trajectories, obtained by adding different noises to the initial safe policy.

At each epoch, we optimize the dynamics models for 1000 steps.

\paragraph{Barrier certificate $\Lya$}
The barrier certificate is parametrized by a feed-forward neural network with ReLU activation and two hidden layers, each of which contains 256 hidden units. The coefficient $\lambda$ in \Cref{eq:L-loss} is set to 0.001.

\paragraph{Collecting data.}
In \env{Upright}, \env{Tilt} and \env{Move}, the Line 3 in \Cref{alg:paradigm} collects a single episode. In \env{Swing}, the Line 3 collects six episodes, two of which are from \Cref{alg:safe-exploration} with a uniform random policy, another two are from the current policy, and the remaining two are from the current policy but with more noises. In \Cref{alg:safe-exploration}, we first draw $n = 100$ Gaussian samples $\zeta_i \sim \mathcal{N}(0, I)$, and the sampled actions are $a_i = \tanh(\mu_\Piparam(s) + \zeta_i \sigma_\Piparam(s))$, where $\sigma_\Piparam(s)$ and $\mu_\Piparam(s)$ are the outputs of the exploration policy $\piexpl$.

\subsection{Baselines}
\label{sec:baselines}

\paragraph{RecoveryRL} We use the code in \url{https://github.com/abalakrishna123/recovery-rl}.
We remark that when running experiments in Recovery RL, we do not add the violation penalty for an unsafe trajectory.
We set $\epsilon_\text{risk} = 0.5$ (chosen from $[0.1, 0.3, 0.7, 0.7]$) and discount factor $\gamma_\text{risk} = 0.6$ (chosen from $[0.8, 0.7, 0.6, 0.5]$).
The offline dataset $\mathcal{D}_\text{offline}$, which is used to pretrain the $Q^\pi_\text{risk}$, contains 20K transitions from a random policy and another 20K transitions from the initial (safe) policy used by CRABS.
The violations in the offline dataset is \textbf{not} counted when plotting.

Unfortunately, with chosen hyperparameters, we do not observe reasonable high reward from the policy, but we do observe that after around 400 episodes, RecoveryRL visits high reward (15-20) region in the \env{Swing} task and there are few violations since then.

\paragraph{SAC} We implement SAC ourselves with learned temperature $\alpha$, which we hypothesize is the reason of it superior performance over RecoveryRL and SQRL. The violation penalty is chosen to be 30 from $[3, 10, 30, 100]$ by tuning in the \env{Swing} and \env{Move} task. We found out that with violation penalty being 100, SAC has slightly fewer violations (around 167), but the total reward can be quite low (< 2) after $10^6$ samples, so we choose to show the result of violation penalty being 30.

\paragraph{SQRL} We use code provided by RecoveryRL with the same offline data and hyperparameters. However, we found out that the $\nu$ parameter (that is, the Lagrangian multiplier) is very important and tune it by choosing the optimal one from $[3, 10, 30, 100, 300]$ in \env{Swing}. The optimal $\nu$ is the same as that for SAC, which is 30.
As SQRL and RecoveryRL use a fixed temperature for SAC, we find it suboptimal in some cases, e.g., for \env{Swing}.

\paragraph{CPO}
We use the code in \url{https://github.com/jachiam/cpo}.
To make CPO more sample efficient and easier to compare, we reduce the batch size from 50000 to 5000 (for \env{Move} and \env{Tilt}) or 1000 (for \env{Tilt} and \env{Upright}). We tune the step size in $[0.02, 0.05, 0.005]$ but do not find substantial difference, while tuning the batch size can significantly reduce its sample efficiency, although it is still sample-inefficient.

\section{Metropolis-Adjusted Langevin Algorithm (MALA)}
\label{sec:mala}

Given a probability density function $p$ on $\RR^d$, Metropolis-Adjusted Langevin Algorithm (MALA) obtains random samples $x \sim p$ when direct sampling is difficult. it is based on Metropolis-Hastings algorithm which generates a sequence of samples $\{x_{t}\}_{t}$. Metropolis-Hastings algorithm requires a \emph{proposal distribution} $q(x'|x)$. At step $t \geq 0$, Metropolis-Hastings algorithm generates a new sample $\hat x_{t+1} \sim q(\cdot | x_t)$ and accept it with probability
$$
\alpha(x \to x') = \min\left( 1, \frac{p(x') q(x | x')}{p(x) q(x' | x)} \right).
$$
If the sample $\hat x_{t+1}$ is accepted, we set $x_{t+1} = \hat x_{t+1}$; Otherwise the old sample $x_t$ is used: $x_{t+1} = x_t$.
MALA considers a special proposal function $q_\tau(x' | x) = \mathcal{N}(x + \tau \nabla p(x), 2 \tau I_d)$. See \Cref{alg:mala} for the pseudocode.

\begin{algorithm}[tb]
    \begin{algorithmic}[1]
        \REQUIRE A probability density function $p$ and a step size $\tau$.
        \STATE Initialize $x_0$ arbitrarily.
        \FOR{$t$ from 0 to $\infty$}
            \STATE Draw $\zeta_t \sim \mathcal{N}(0, I_d)$.
            \STATE Set $\hat x_{t+1} = x_t + \tau \nabla \log p(X_t) + \sqrt{2 \tau} \zeta_t$.
            \STATE Draw $u_t \sim \text{Uniform}[0, 1]$.
            \IF{$u_t \geq \alpha(x_t \to \hat x_{t+1})$}
                \STATE Set $x_{t+1} = \hat x_{t+1}$.
            \ELSE
                \STATE Set $x_{t+1} = x_t$.
            \ENDIF
        \ENDFOR
    \end{algorithmic}
    \caption{Metropolis-Adjusted Langevin Algorithm (MALA)}
    \label{alg:mala}
\end{algorithm}

For our purpose, as we seek to compute $C^*(\Lya_\Lparam, U, \policy_\Piparam)$, we maintain $m = 10^4$ sequences of samples $\{\{s^{(i)}_{t}\}_t\}_{i \in [m]}$. Recall that $C^*$ involves a constrained optimization problem:
$$
C^*(\Lya_{\Lparam}, \U, \policy_\Piparam) := \max_{s: \Lya_\Lparam(s) \leq 1} \U(s, \policy_\Piparam(s), \Lya_\Lparam),
$$
so for each $i \in [m]$, the sequence $\left\{s^{(i)}_t \right\}_t$ follows the \Cref{alg:mala} to sample $s \sim \exp(\lambda_1 U(s, \policy_\Piparam(s), \Lya_\Lparam) - \lambda_2 \mathbb{I}_{s \in \Ch})$ with $\lambda_1 = 30, \lambda_2 = 1000$. The step size $\tau$ is chosen such that the acceptance rate is approximately 0.6. In practice, when $s_t^{(i)} \not\in \Ch$, we do not use MALA, but use gradient descent to project it back to the set $\Ch$.

\end{document}